\def\year{2017}\relax

\documentclass[letterpaper]{article}
\usepackage{aaai17}

\nocopyright

\usepackage{times}
\usepackage{relsize} % relative font sizes (e.g. \smaller). must precede ACL style
\usepackage[colorlinks=true,linkcolor=black,citecolor=black,filecolor=black,urlcolor=black]{hyperref}

\usepackage{microtype}

\usepackage[boxed]{algorithm2e}

\usepackage[small,bf,skip=5pt]{caption}
\usepackage{sidecap} % side captions
\usepackage{rotating}	% sideways

\usepackage{titlesec}
\titleformat*{\subparagraph}{\itshape}
\titlespacing{\subparagraph}{%
  1em}{%              left margin
  0pt}{% space before (vertical)
  1em}%               space after (horizontal)

\usepackage{lingmacros}
\usepackage[shortlabels]{enumitem} % customizable lists
\setlist{nolistsep}

\usepackage{verbatim}

\usepackage{natbib}

\usepackage{xspace}
\usepackage{xparse} % for fancy custom macros (\NewDocumentEnvironment, etc.)

\usepackage{textcomp}
\usepackage{framed}

\usepackage{listings}

\usepackage{amssymb}	%amsfonts,eucal,amsbsy,amsthm,amsopn
\usepackage{amsmath}

\usepackage{mathptmx}	% txfonts
\usepackage[scaled=.8]{beramono}
\usepackage[scaled=.85]{helvet}
\usepackage[T1]{fontenc}
\usepackage[utf8x]{inputenc}

\usepackage{MnSymbol}	% must be after mathptmx
\usepackage{latexsym}

\makeatletter
\DeclareTextCommandDefault{\fourdots}{%
.\kern\fontdimen3\font
.\kern\fontdimen3\font
.\kern\fontdimen3\font
.\kern\fontdimen3\font}
\makeatother

\usepackage{array}
\usepackage{longtable}
\usepackage{multirow}
\usepackage{booktabs} % pretty tables
\usepackage{multicol}
\usepackage{footnote} % for savenotes environment
\newcolumntype{H}{>{\setbox0=\hbox\bgroup}c<{\egroup}@{}} % hidden column

\usepackage{url}
\usepackage[usenames,dvipsnames,svgnames,table]{xcolor}

\usepackage[normalem]{ulem} % \uline
\usepackage{colortbl}
\usepackage{graphicx}
\usepackage{subcaption}
\usepackage{tikz}
\usetikzlibrary{arrows,positioning,calc}

\usepackage{nameref}
\usepackage{cleveref}

\crefformat{part}{\S#2#1#3}
\crefformat{chapter}{\S#2#1#3}
\crefformat{section}{\S#2#1#3}
\crefformat{subsection}{\S#2#1#3}
\crefformat{subsubsection}{\S#2#1#3}
\crefformat{paragraph}{\P#2#1#3}
\crefformat{subparagraph}{\P#2#1#3}
\crefmultiformat{section}{\S#2#1#3}{ and~\S#2#1#3}{, \S#2#1#3}{, and~\S#2#1#3}
\crefmultiformat{subsection}{\S#2#1#3}{ and~\S#2#1#3}{, \S#2#1#3}{, and~\S#2#1#3}
\crefmultiformat{subsubsection}{\S#2#1#3}{ and~\S#2#1#3}{, \S#2#1#3}{, and~\S#2#1#3}
\crefmultiformat{paragraph}{\P\P#2#1#3}{ and~#2#1#3}{, #2#1#3}{, and~#2#1#3}
\crefmultiformat{subparagraph}{\P\P#2#1#3}{ and~#2#1#3}{, #2#1#3}{, and~#2#1#3}
\crefrangeformat{section}{\mbox{\S\S#3#1#4--#5#2#6}}
\crefrangeformat{subsection}{\mbox{\S\S#3#1#4--#5#2#6}}
\crefrangeformat{subsubsection}{\mbox{\S\S#3#1#4--#5#2#6}}
\crefrangeformat{paragraph}{\mbox{\P\P#3#1#4--#5#2#6}}
\crefrangeformat{subparagraph}{\mbox{\P\P#3#1#4--#5#2#6}}
\crefname{part}{Part}{Parts}
\Crefname{part}{Part}{Parts}
\crefname{chapter}{ch.}{ch.}
\Crefname{chapter}{Ch.}{Ch.}
\crefname{figure}{figure}{figures}
\crefname{subfigure}{figure}{figures}
\crefname{appsec}{appendix}{appendices}
\Crefname{appsec}{Appendix}{Appendices}
\crefname{algocf}{algorithm}{algorithms}
\Crefname{algocf}{Algorithm}{Algorithms}
\crefname{enums}{example}{examples}
\Crefname{enums}{Example}{Examples}
\crefname{enumsi}{example}{examples}
\Crefname{enumsi}{Example}{Examples}
\crefname{}{example}{examples} % lingmacros \toplabel has no internal name for the kind of label
\Crefname{}{Example}{Examples}
\crefformat{enums}{(#2#1#3)}
\crefformat{enumsi}{(#2#1#3)}
\crefformat{}{(#2#1#3)}
\crefrangeformat{enums}{\mbox{(#3#1#4--#5#2#6)}}
\crefrangeformat{enumsi}{\mbox{(#3#1#4--#5#2#6)}}
\crefmultiformat{enumsi}{(#2#1#3}{, #2#1#3)}{, #2#1#3}{, #2#1#3)}
\crefrangemultiformat{enumsi}{(#3#1#4--#5#2#6}{, #3#1#4--#5#2#6)}{, #3#1#4--#5#2#6}{, #3#1#4--#5#2#6)}

\ifx\creflastconjunction\undefined%
\newcommand{\creflastconjunction}{, and\nobreakspace} % Oxford comma for lists
\else%
\renewcommand{\creflastconjunction}{, and\nobreakspace} % Oxford comma for lists
\fi%

\newcommand*{\Fullref}[1]{\hyperref[{#1}]{\Cref*{#1}: \nameref*{#1}}}
\newcommand*{\fullref}[1]{\hyperref[{#1}]{\cref*{#1}: \nameref{#1}}}
\NewDocumentEnvironment{itmize}{}{\begin{itemize}[noitemsep]}{\end{itemize}}
\NewDocumentEnvironment{enumrate}{}{\begin{enumerate}[noitemsep]}{\end{enumerate}}
\let\Item\item
\renewcommand\enddescription{\endlist\global\let\item\Item}
\NewDocumentEnvironment{describe}{}{\renewcommand\item[1][]{\Item \textbf{##1:} }\begin{itemize}}{\end{itemize}}
\NewDocumentEnvironment{edescribe}{}{\renewcommand\item[1][]{\Item \textbf{##1:} }\begin{enumerate}}{\end{enumerate}}

\usepackage{color}
\usepackage{bm}
\definecolor{orange}{rgb}{1,0.5,0}
\definecolor{mdgreen}{rgb}{0,0.6,0}
\definecolor{mdblue}{rgb}{0,0,0.7}
\definecolor{dkblue}{rgb}{0,0,0.5}
\definecolor{dkgray}{rgb}{0.3,0.3,0.3}
\definecolor{slate}{rgb}{0.25,0.25,0.4}
\definecolor{gray}{rgb}{0.5,0.5,0.5}
\definecolor{ltgray}{rgb}{0.7,0.7,0.7}
\definecolor{purple}{rgb}{0.7,0,1.0}
\definecolor{lavender}{rgb}{0.65,0.55,1.0}

\makeatletter
\renewcommand{\paragraph}{%
  \@startsection{paragraph}{4}%
  {\z@}{.2ex \@plus 1ex \@minus .2ex}{-.7em}%
  {\bfseries}%
}
\makeatother

\newcommand{\p}[1]{\textbf{\textsf{#1}}} % preposition type
\newcommand{\lbl}[1]{\textsc{#1}} % class label
\newcommand{\sst}[1]{\lbl{#1}} % supersense tag label
\newcommand{\psst}[1]{\textcolor{mdgreen}{\sst{#1}}} % preposition supersense tag label
\newcommand{\rf}[2]{\psst{#1}$\leadsto$\psst{#2}}
\newcommand{\rff}[3]{\psst{#1}$\leadsto$\psst{#2}$\leadsto$\psst{#3}}

\newcommand{\finalversion}[1]{#1}

\newcommand{\nonanonversion}[1]{}
\newcommand{\shortversion}[1]{}

\newcommand{\longversion}[1]{#1} % ...if only there were more space...
\newcommand{\subversion}[1]{} % for the submission version only
\newcommand{\ignore}[1]{}

\begin{document}

%\title{Bipartite Analysis for Broad-Coverage Semantic Annotation of Adpositions}%\ab{wasn't it tripartite?}
\title{Coping with Construals in Broad-Coverage Semantic Annotation of Adpositions}
\author{
Jena D. Hwang \and
Archna Bhatia \\
	IHMC \\
    {\tt \{jhwang,abhatia\}@ihmc.us} \And
Na-Rae Han \\
	University of Pittsburgh \\
    {\tt naraehan@pitt.edu} \AND 
Tim O'Gorman \\
	University of Colorado Boulder \\
    {\tt timothy.ogorman@colorado.edu} \And
Vivek Srikumar \\
	University of Utah \\
    {\tt svivek@cs.utah.edu} \And
Nathan Schneider \\
	Georgetown University \\
    {\tt nathan.schneider@georgetown.edu} \\
% \textsuperscript{1}The Florida Institute for Human and Machine Cognition (IHMC)\\
% \textsuperscript{2}University of Pennsylvania\\
% \textsuperscript{3}University of Colorado Boulder\\
% \textsuperscript{4}University of Utah\\
% \textsuperscript{5}Georgetown University\\
% \{jhwang,abhatia\}@ihmc.us, naraehan@pitt.edu, timothy.ogorman@colorado.edu,\\ svivek@cs.utah.edu, nathan.schneider@georgeown.edu\\
}
\maketitle
\begin{abstract} % emphasize NLU goals?
We consider the semantics of prepositions, 
revisiting a broad-coverage annotation scheme used for annotating 
%that we have used to annotate 
all 4,250 preposition tokens in a 55,000 word corpus of English. 
Attempts to apply the scheme to adpositions and case markers in other languages, 
as well as some problematic cases in English, have led us to reconsider the assumption
that a preposition’s lexical contribution is equivalent to the role/relation 
that it mediates. Our proposal is to embrace the potential for 
%productive metaphor and 
\textbf{construal} in adposition use, expressing such phenomena directly 
at the token level to manage complexity and avoid sense proliferation. 
We suggest a framework to represent both the scene role and the adposition's lexical function 
so they can be annotated at scale---supporting automatic, statistical processing 
of domain-general language---and sketch how this representation 
would inform a constructional analysis.
%We sketch an analysis of how the role semantics and function semantics can be 
%simultaneously evoked in a \textbf{bipartite analysis}.
%This endeavor serves as a case study of balancing between the 
%theoretical flexibility offered by Construction Grammar and 
%the practical need for parsimony in order to achieve corpus coverage.
\end{abstract}

\section{Introduction}

A consequence of the vast expressiveness of human language is that 
natural language understanding (NLU) cannot scale to \emph{general language} input 
unless it is willing to make some compromises for the sake of practicality 
and robustness. One such compromise made in most state-of-the-art 
natural language processing technologies (e.g., syntactic parsing) is that 
the computational model of language is not a complete model 
of human grammatical knowledge, but rather, a set of soft preferences 
derived by statistical learning algorithms from large human-annotated datasets. 
Thus, a strategy for advancing the state-of-the-art in NLU 
%is to focus linguists' descriptive effort not on hand-built grammar rules, but on 
% jena: i see you added annotation scheme below, but i've the feeling that much like what
% the reviewer said much of what "rules" mean here is going to be somewhat elusive to
% pure linguistic readers. And besides, there is no reason that a classifier shouldn't 
% be accompanied or be aided by rules, either, so it wouldn't be too grievous to leave
% the first bit out
is to focus linguists' descriptive effort on
\emph{annotation schemes} and \emph{datasets}: 
annotating corpora with semantic information, for instance, 
so that formal cues (denotational or contextual) can be automatically associated with meaning representations.
A representation will necessarily be limited in the level of detail it provides (its \emph{granularity}) and/or the range of linguistic expressions that it is prepared to describe (its \emph{coverage}). 
The principles of Construction Grammar can inform corpus annotations 
even if they fall short of full-fledged constructional parses.

Mindful of this granularity--coverage tradeoff, we have sought to develop a scheme that will be of practical value for broad-coverage human annotation, and therefore domain-general NLU, for a particular set of lexicogrammatical markers: \textbf{prepositions} in English, and more generally, \textbf{adpositions} and \textbf{case markers} across languages. Forming a relatively closed class, these markers are incredibly versatile, and therefore exceptionally challenging to characterize semantically, let alone disambiguate automatically (\cref{sec:lit}).

As a first step, we describe \textbf{preposition supersenses}, which target a coarse level of granularity and support comprehensive coverage of types and tokens in English.
However, in attempting to generalize this approach to other languages, we uncovered a major weakness: it does not distinguish the contribution of the preposition itself, i.e., what the adposition \textbf{codes} for, from the semantic role or relation that the adposition mediates and that a predicate or scene \textbf{calls} for; and as a result, the label that would be most appropriate is underdetermined for many tokens  %This has caused persistent issues in English supersense annotation and its adaptation to other languages
(\cref{sec:problems}).
In our view, the mismatch can be understood through the lens of \textbf{construal} and should be made explicit, leveraging the principles of Construction Grammar (\cref{sec:construal}). 
\cref{sec:applying_bipartite} surveys some of the phenomena that our new analysis addresses; %\cref{sec:problems} 
\cref{sec:challenges} discusses the tradeoffs inherent in the proposed approach.
Finally, we sketch how our proposal would fit into a compositional constructional analysis of adpositional phrases %(\cref{sec:cxg}).
(\cref{sec:cxg_discussion}).

%\jh{paper roadmap: (1) background, (2) Bipartite analysis....}
% \psst{While we are building corpora with semantic labels rather than a full-fledged construction grammar, our analysis informs how such a grammar would represent the compositionality of predicates, prepositions, and objects. In \cref{sec:cxg}, we sketch how a construction grammar analysis could capture the function and role at different levels of structure, similar to constructional treatments of metaphor in previous work.}

%\nss{what are the broader implications of this study?}

%%%%%%%%%%%%%

\section{Approaches to Prepositional Polysemy}\label{sec:lit}
%In this section, we discuss prior approaches to prepositional polysemy.
\begin{figure*}\centering
\includegraphics[width=.7\textwidth]{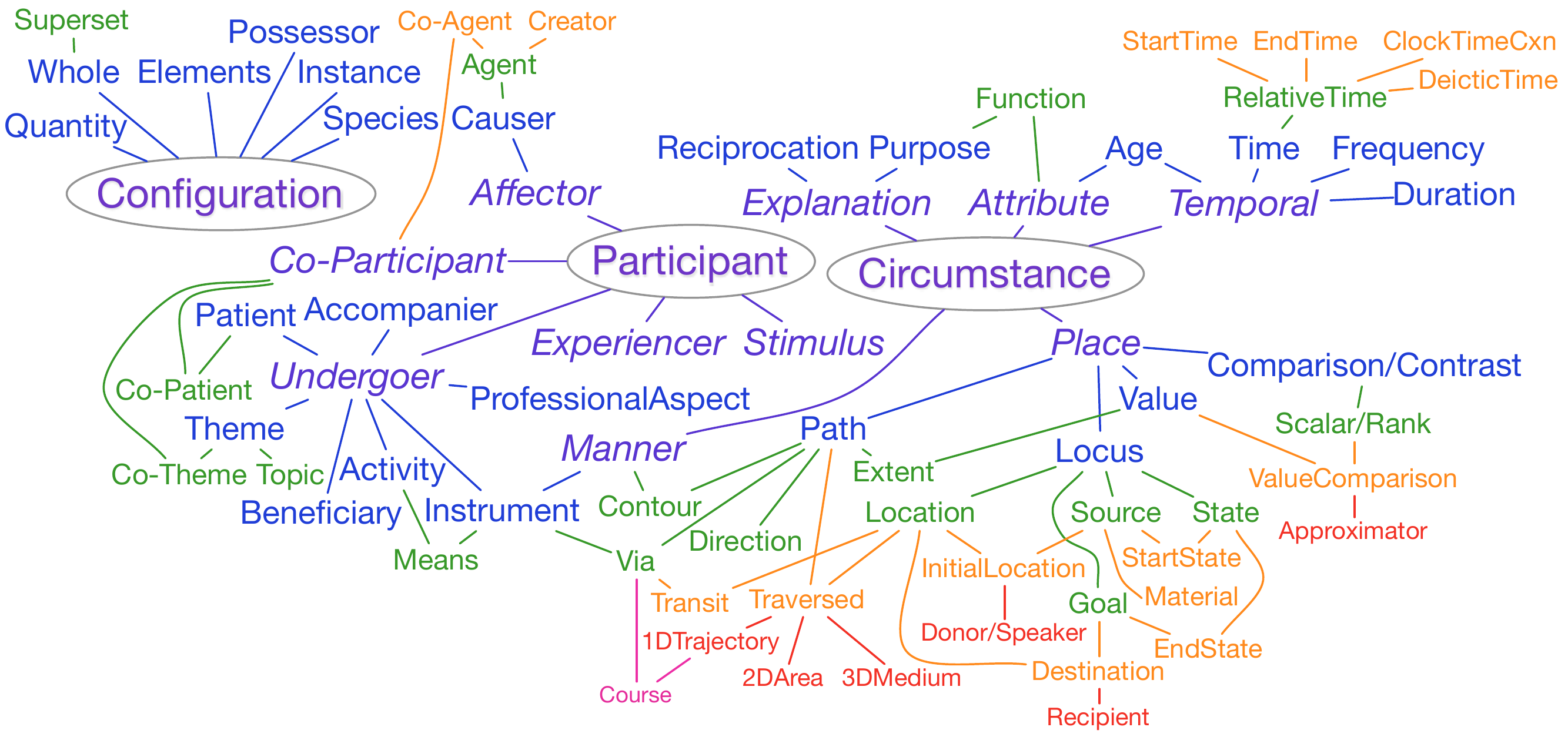}
\caption{Preposition supersense hierarchy \citep[from][]{schneider-16}. Top-level categories are circled 
and subcategories radiate outward.}
\label{fig:hierarchy}
\end{figure*}

%\subsection{}

The most frequent English prepositions are extraordinarily polysemous. For example, the preposition \p{at} expresses different information in each of the following usages:

\eenumsentence{\label{ex:at}
\itemsep0em 
\item \label{ex:2} The coffee shop is \p{at} 123 Main St. (\psst{Location})
\item \label{ex:1} We met him \p{at} 7pm. (\psst{Time})
%\item \label{ex:3} We met him \p{at} a conference.
\item \label{ex:4} Suddenly, everyone pointed \p{at} him. (\psst{Goal}) % was 'looked at'. PrepWiki currently says Goal, but I would say Stimulus ~> Goal
\item \label{ex:5} She laughed \p{at} my acting. (\psst{Stimulus})
\item \label{ex:6} They were robbed \p{at} gunpoint. (\psst{Instrument})}
NLU systems, when confronted with a new instance of \p{at}, must determine whether it marks an entity or scene's location, time, instrument, or something else.

As lexical classes go, prepositions are something of a red-headed stepchild in the linguistics literature. Most of the semantics literature on prepositions has revolved around how they categorize space and time  \citep[e.g.,][]{herskovits-86,verkuyl-92,bowerman-01}. However, there have been a couple of lines of work addressing preposition semantics broadly. %\nss{\citep{jones-87}}
In cognitive linguistics, studies have examined abstract as well as concrete uses of English prepositions \citep[e.g.,][]{dirven-93,
%dirven-95,
lindstromberg-10}. Notably, the polysemy of \p{over} and other prepositions has been explained in terms of sense networks encompassing core senses and motivated extensions \citep{brugman-81,lakoff-87,dewell-94,tyler-01,tyler-03}. 
The Preposition Project \citep[TPP;][]{litkowski-05} broke ground in stimulating computational work on fine-grained word sense disambiguation of English prepositions \citep{litkowski-05,ye-07,tratz-09,dahlmeier-09}.
Typologists, meanwhile, have developed \emph{semantic maps} of functions, where the nearness of two functions reflects their tendency to fall under the same adposition or case marker in many languages \citep{haspelmath-03,walchli-10}.

%Like much of the computational literature, our goal is to facilitate semantic disambiguation. 
%\subsection{Preposition Supersenses}\label{sec:prepsupersenses}

\paragraph{Preposition supersenses.} Following \citet{srikumar-13}, we sought coarse-grained semantic categories of prepositions as a broader-coverage alternative to fine-grained senses. 
Because we want our labels to generalize across languages, we use categories similar to those appearing in semantic maps (\psst{Location}, \psst{Recipient}, etc.)\ rather than lexicalized senses. 
%Our labels are organized into a taxonomy, as opposed to a network reflecting category extension or typological cooccurrence.
We identified a set of such categories through extensive deliberation involving the use of dictionaries, corpora and pilot annotation experiments \citep{schneider-15}.
We call these categories \textbf{supersenses} to emphasize their similarity to coarse-grained classifications of nouns and verbs that go by that name \citep{ciaramita-06,schneider-12}. 
The \p{at} examples in \cref{ex:at} are accompanied by the appropriate supersenses from our scheme. 
Most supersenses resemble thematic roles, in the tradition begun by \citet{fillmore-68}; a few others are needed to describe preposition-marked relations between entities.
There are multiple English prepositions per supersense; e.g., %for example, 
``\p{in} the city'' and ``\p{on} the table'' would join ``\p{at} 123 Main St.'' in being labeled as \psst{Location}s. 
%Supersenses generally capture the prototypical semantics expressed by the preposition. \jh{added}
We understand the supersenses as prototype-based categories, and in some cases use heuristics like paraphraseability (``in order to'' for \psst{Purpose}) 
and WH-question words (``Why?'' for \psst{Purpose} and \psst{Explanation}) to help determine which tokens are instances of the category.%\nss{OK?}\ab{sounds ok to me}

The 75~supersenses are organized in a taxonomy based on that of VerbNet \citep{bonial-11}, with \psst{Participant}, \psst{Circumstance}, and \psst{Configuration} at the top level.\footnote{These loosely correspond to event arguments, adjuncts, and adnominal complements, respectively. 
%But these are merely prototypes; w
However, we do not make any claims with regard to coreness or the argument/adjunct distinction, 
as there are many phenomena that do not conform to either of the prototypes for argument and adjunct \citep[for a review of the literature on the argument/adjunct distinction, see][]{hwang-11}. 
We are also not convinced that a firm distinction between lexical and nonlexical/functional adpositions \citep{rauh-93} 
can be established, though the relevance of this distinction in the context of the bipartite construal approach merits further investigation.} 
The taxonomy uses multiple inheritance to account for subcategories which are considered to include properties of multiple supercategories.
The full hierarchy appears in \cref{fig:hierarchy}.

Our approach to preposition annotation is \emph{comprehensive}, i.e., every token of every preposition type is given a supersense label. 
We applied the supersenses to annotate a 55,000 word corpus of online reviews in English, covering all 4,250 preposition tokens \citep{schneider-16}.
For each token, annotators chose a single label from the inventory. 
This is not an easy task, but with documentation of many examples in a lexical resource, \textbf{PrepWiki},\footnote{\url{http://tiny.cc/prepwiki}}
trained university students were able to achieve reasonable levels of inter-annotator agreement.
Every token was initially labeled by at least two independent annotators, 
and differences were adjudicated by experts.

%hierarchy\\
%prepwiki\\
%practical needs of annotation process\\

\section{Problems with Preposition Supersenses}\label{sec:problems}

While the above approach worked reasonably well for most English tokens, 
a few persistent issues arising in English and other languages have led us to revisit 
%the\nss{application of the?}\ab{I think it's more like: ...a few persistent issues arising in English and attempts to apply the hierarchy to other languages have led us to revisit the } supersense taxonomy as well as 
fundamental assumptions about what it means to semantically label an adposition.

\subsection{Semantic Overlap} \label{sec:overlap}

In our original English annotation \citep{schneider-16}, a few phenomena 
caused us much hand-wringing---not because there was no appropriate supersense, but because \emph{multiple} supersenses seemed to fit. For example, we found that \psst{Topic} and \psst{Stimulus} could compete for semantic territory.

\cref{ex:about-topic}~evinces related usages of \p{about} with different governors:

\eenumsentence{\label{ex:about-topic}
\itemsep0em 
\item\label{ex:book-about} I read [a book \p{about} the strategy].
\item\label{ex:read-about} I read \p{about} the strategy.
\item\label{ex:know-about} I knew \p{about} the strategy.
\item\label{ex:care-about} I cared \p{about} the strategy.
%    [I didn't care FOR the strategy.: the prep disambiguates the sense of the pred?]
}

%All four usages could reasonably be labeled as \psst{Topic}. This is because the \p{about}-PP indicates what is communicated \cref{ex:book-about,ex:read-about}, known \cref{ex:know-about}, or judged, which implies it is known \cref{ex:care-about}.
%\nss{need to rethink this: by this logic, in ``I love the book'', is the book a topic?}\ab{yes, I am not sure about 2d!}\jh{I don't know, to me TOPIC for cared about works. Because of care you get the reading that this is an emotional process and the strategy is the trigger of that emotion (i.e. I love/hate this stategy). But by using "about" along with care instead of "for" it feels like you are zeroing in on the more heady (thought-related) aspect of caring. I cared about the strategy (I thought about it, I consternated over it, and I made it my business). }\ab{hmm I guess that is the point, "about" introduces this construal when we focus on its contribution. I think its just that the emotional aspect is very strong in this case (2d) instead of the content being introduced by "about". Now that I think about it, (2d) does seem different from "I love the book", they seemed similar because of the emotional aspect being common, but they are different too in that "about" does introduce that construal in (2d) which we do not have in this "love the book" example, that is why TOPIC reading is not accessible for this example.}\jh{I think we'd all agree love the book does not have topic reading.} 
%All are forms of information content.

The first three usages could reasonably be labeled as \psst{Topic}. This is because the \p{about}-PP indicates what is communicated \cref{ex:book-about,ex:read-about} and known \cref{ex:know-about}. The fourth example \cref{ex:care-about}, however, presents an overlap in its interpretation. On the one hand, 
%However, \cref{ex:care-about} presents a problem because 
traditional thematic role inventories include the category \psst{Stimulus} for something that prompts a perceptual or emotional experience, as in \cref{ex:afraid-of}. 

\enumsentence{\label{ex:afraid-of}
I was afraid \p{of} the strategy.}

Surely, \emph{cared} in \cref{ex:care-about} describes an emotional state, so \p{about} marks the \psst{Stimulus}.  However, much like examples \cref{ex:book-about,ex:read-about,ex:know-about}, the semantics relating to \psst{Topic} is still very much present in the use of \p{about}, which draws attention to the aspects of the caring process involving thought or judgement. This sits in contrast to the use of \p{for} in ``I cared \p{for} my grandmother,'' %
%\cref{ex:care-for}
where the prepositional choice calls attention to the benefactive aspect of the caring act. 

% \enumsentence{\label{ex:care-for}
% I cared \p{for} my grandmother.}

If we are constrained to one label per argument, where should the line be drawn between \psst{Stimulus} and \psst{Topic} in cases of overlap? In other words, should the semantic representation emphasize the semantic commonality between all of the examples in \cref{ex:about-topic}, or between \cref{ex:care-about} and \cref{ex:afraid-of}?

Observing that annotators were inconsistent on such tokens, we 
%went to great lengths to 
drew a boundary between \psst{Topic} and \psst{Stimulus} in an attempt to force consistency. Below, we instead argue that the idea of construal\slash conceptualization offers a more principled answer; in our new analysis, the \psst{Topic} suggested by \p{about} and the \psst{Stimulus} suggested by \emph{cared} can coexist.
%We can say that in this case, \textbf{the preposition construes the scene}: that is, the preposition casts the trigger of emotional experience as information content.

\subsection{Applying the Supersenses to Other Languages}\label{sec:probOtherLgs}

One of the premises of using unlexicalized supersenses was that the scheme would port well to other languages (as the WordNet noun and verb supersenses have: \citealp[\emph{inter alia}]{picca-08,schneider-12}).
To test this, we have begun applying the existing supersenses to three new languages, namely, Hebrew, Hindi, and Korean. Pilot annotation in these languages has echoed the fundamental problem discussed in the previous section.

Consider the Hindi examples below. In \cref{ex:bipasha}, the experiencer of an emotion is marked with a postposition \p{kaa}, the genitive case marker in Hindi. %Prototypically this  postposition is used to mark possessors. %The supersense label \psst{Experiencer} alone misses out on the semantics of possession originating from the postposition \p{kaa} in \cref{ex:bipasha}. 
The use of \p{kaa} strongly suggests possession (here it is possession of an abstract quality). However, the semantics of the phrase also includes \psst{Experiencer}---thus, it seems inappropriate to choose between \psst{Experiencer} and \psst{Possessor} for this token. (The same problem is encountered with a similar phrase ``the anger \p{of} Bipasha'' in English.) 
There are other ways to attribute anger to Bipasha---e.g., see \cref{ex:counter_bipasha}. Here Bipasha is not construed as a possessor when the postposition \p{kaa} is not used.
%\ab{is this statement sensible to make under the current construal understanding? or should we restate it differently? here i am saying that the postposition's complement is construed as possessor because of the postposition, but probably we cannot say that now since construal involves both the parts?! Just check it once.}\jh{this sounds fine.}\ab{ok} \jh{for the most part there's really nothing in the document, unless i totally missed it that sounds wrong -- except the direct statement. i will look over it again later.}%, a construal that is not present in example \cref{ex:counter_bipasha} where the postposition \emph{kaa} is not used. 

\eenumsentence{\itemsep0em 
\item \label{ex:bipasha} %बिपाशा \textbf{का} गुस्सा 
[Hindi]:  \psst{Experiencer} vs.~\psst{Possessor}\\
bipaashaa     \p{kaa}     gussaa \\
Bipasha         \p{GEN}        anger \\
``Bipasha’s anger''

%\item \label{ex:bipasha2}  
%[Hindi]:  \psst{Possessor} \\
%bipaashaa     \textbf{kaa}     ghar \\
%Bipasha         \p{GEN}        house \\
%``Bipasha’s house''

\item \label{ex:counter_bipasha}
[Hindi]:  \psst{Experiencer}\\
bipaashaa     bahut     gussaa	hui\\
Bipasha         very	angry	became \\
``Bipasha got very angry.''

% \item \label{ex:happiness} %우리\textbf{의}  기쁨 
% [Korean]:  \psst{Experiencer} vs.~\psst{Possessor}\\
% uri-\textbf{ui}     gipum \\
% 1pers.pl-\p{GEN}      happiness \\
% ``our happiness''
}

Our preliminary annotation of Hindi, Korean, and Hebrew 
has suggested that instances of overlap between multiple supersenses 
are fairly frequent. %, introducing new construals that might have been missed at the English annotation stage.
%far more frequent than might have been expected from English.\ab{is it true that these are more frequent in other languages? or is it just that while annotating English, the idea of construal was not yet there so these cases went unnoticed?}\jh{probably. is the edited version better?}

\section{Bipartite Construal Analysis} \label{sec:construal}

Why do ``cared \p{about} the strategy'' in \cref{ex:care-about} and ``anger \p{of} Bipasha'' in \cref{ex:bipasha} above not lend themselves to a single label?
%Examples such as \emph{cared \p{about} the strategy} and \emph{anger \p{of} Bipasha} do not lend themselves to a single label. 
These seem to be symptoms of the fact that no English preposition prototypically marks \psst{Experiencer} or \psst{Stimulus} roles, though from the perspective of the predicates, such roles are thought to be important generalizations in characterizing events of perception and emotion.
In essence, there is an apparent mismatch between the roles that the verb \emph{care} or the noun \emph{anger} calls for, and the functions that English prepositions prototypically code for. 
While \p{about} prototypically codes for \psst{Topic} and \p{of} prototypically codes for \psst{Possessor}, there is no preposition that ``naturally'' codes for \psst{Experiencer} or \psst{Stimulus} in the same way.
Thus, if a predicate marks an \psst{Experiencer} or \psst{Stimulus} with a preposition, 
the preposition will contribute something new to the conceptualization of the scene being described. With ``cared \p{about} the strategy,'' it is \psst{Topic}-ness that the preposition brings to the table; with ``anger \p{of} Bipasha,'' it is the conceptualization of anger as an attribute that somebody possesses.

%% Going ahead and removing this
%\nss{Can we remove this paragraph?}\jh{ya it's redundant.}As the multiple examples presented here from across languages will illustrate, these issues become starker when attempting semantic annotation of adpositions cross-linguistically.  Much like taking a picture of a scene from different viewpoints will result in different renderings, analyzing a scene (i.e., a situation or an event) from an adposition’s or predicate's viewpoint may give rise to slightly different conceptualizations of the scene. We find that these conceptualizations of scenes with regards to prepositions are handled differently by different languages, each language conceptualizing a situation a little differently from another language as evidenced by the choices of adpositions or case markers.

Thus, we turn to the theories in Cognitive Semantics to define the phenomenon of \textbf{construal} as a means of understanding the contributions that are emerging from the adpositions with respect to the expressed event or situation. Then, we turn to the guiding principles of Construction Grammar to develop a method called \textbf{bipartite analysis} in order to handle the problem posed by construals and to resolve the apparent semantic overlap which is pervasive across languages.

\subsection{Construal}

%Adopting the Notion of Construal \\ from Cognitive Semantics}

The world is not neatly organized into bits of information that map directly to linguistic symbols. 
%Cognitive linguists distinguish between \emph{conceptual structure}, which is taken to include universal aspects of cognition, 
%and \emph{semantic structure} reflecting 
Rather, linguistic meaning reflects the priorities and categorizations of particular expressions in a language \citep[ch.~3]{
%jackendoff-92,pinker-89,rappaport-98,
langacker-98,jackendoff-02,croft-04}. %\nss{maybe also cite Talmy, Lakoff, Johnson.} %\jh{can keep one jackendoff (he specifically talks about conceptual structbure to semantics mapping) and replace the rest; especially if we can reuse citations} 
Much like pictures of a scene from different viewpoints will result in different renderings, a real-world situation being described will ``look'' different depending on the linguistic choices made by a speaker. This includes within-language choices: e.g., the choice of ``John sold Mary a book'' vs.\ ``John sold a book to Mary'' vs.\ ``Mary bought a book from John.'' In the process called \textbf{construal}
%of semantics signaled by the preposition from the role expected by the scene,
(a.k.a.~\textbf{conceptualization}), a speaker ``packages'' ideas for linguistic expression 
in a way that foregrounds certain elements of a situation while backgrounding others. %, making the leap from conceptual structure to semantic structure. 
% This process is potentially lossy: the affordances of the language (grammar and vocabulary; linearity of the signal) 
% may not perfectly match the idea to be expressed, and therefore, some nuances of the original idea may be lost, or other nuances added, as it enters semantic structure to be realized linguistically. 

We propose to incorporate this notion of construal in adposition supersense annotation.
%to refer to the nuances of meaning that emerge when a particular preposition is chosen over another given the described scene. 
%We adopt this notion of construal to supersense annotation to refer to the semantics signaled by the adposition given the role expected by the scene.\nss{awkward}
We use the term \textbf{scene} to refer to events or situations in which an adpositional phrase plays a role.
(We do not formalize the full scene, but %we 
assume its roles can be characterized 
with supersense labels from \cref{fig:hierarchy}.)
Contrast the use of the prepositions \p{by} and \p{of} %sentences 
in %example 
\cref{ex:puccini}: 
\eenumsentence{ \label{ex:puccini}
\itemsep0em 
\item \label{ex:puccini-by} The festival focuses on the works \p{by} Puccini. 
\item \label{ex:puccini-of} He was an expert on the works \p{of} Puccini. 
}
While both prepositional phrases indicate works created by the operatic composer Puccini (i.e., \psst{Creator}), the different choices of preposition reflect different construals: \p{by} %in example \cref{ex:puccini-by} 
highlights the agency of Puccini, %. This construal, however, is missing from example \cref{ex:puccini-of}. The construal available in \cref{ex:puccini-of} 
whereas \p{of} construes Puccini as the source of his composition.
Thus, ``works \p{by} Puccini'' and ``works \p{of} Puccini'' are paraphrases, 
but present subtly different portrayals of the relationship between Puccini and his works. 
In other words, these paraphrases are not identical in meaning 
because the preposition carries with it different nuances of construal.
In this paper, we focus on differences in construal manifested in different adposition choices, 
and the possibility that an adposition construal complements the construal of a scene and its roles (as evoked by the governing head or predicate).

For instances  like ``I read \p{about} the strategy'' in \cref{ex:read-about} that were generally unproblematic for annotation under the original preposition guidelines, the semantics of the adposition and the semantic role assigned by the predicate are congruent. However, for examples like ``cared \p{about} the strategy'' in \cref{ex:care-about} and ``anger \p{of} Bipasha'' in \cref{ex:bipasha}, we say that the adposition construes the role as something other than what the scene specifies. Competition between different adposition construals accounts for many of the alternations that are near-paraphrases, but potentially involve slightly different nuances of meaning (e.g., ``talk \p{to} someone'' vs.~``talk \p{with} someone''; ``angry \p{at} someone'' vs.~``angry \p{with} someone''). 

Thus, the notion of construal challenges \citeauthor{schneider-15}'s (\citeyear{schneider-15,schneider-16}) original conception that each supersense reflects the semantic role assigned by its governing predicate (i.e.~verbal or event nominal predicate), %---for example, verbal or event nominal predicates---
%to their adpositional arguments or adjuncts, 
and that a single supersense label can be assigned to each adposition token. Rather than trying to ignore these construals to favor a single-label approach, or possibly create new labels to capture the meaning distinctions that construals impose on semantic roles, we adopt an approach that gives us the flexibility to deal with both the semantics coming from the scene as well as the construals imposed by the adpositional choice.

% The nuanced differences in conceptualization brought on by the prepositional choice threatened a proliferation in supersense labels.
% %resulted in  proliferation of supersense labels.\nss{example}\ab{I think proliferation would have been due to prepositions seeming to have more specific or more general content than semantic roles, other than that I'm not sure why mapping to semantic roles would result in proliferation of labels, but you guys developed the hierarchy, do you remember when you realized there was a need for a new label, what was it? and which labels did you need to add?} 
% Additionally, while annotating prepositions with supersense labels, there were problematic tokens where multiple supersenses seemed to apply, forcing an arbitrary decision between these or the added complexity of a new category. 

% %Hence, taking adpositions' supersense labels to be equivalent to semantic roles results in a problem since a single role that is geared towards the representation of semantic relationship of the prepositional phrase to the predicate does not satisfactorily account for the differing construals that arise from the prepositional choice in the sentence. 

\subsection{Formulating a Bipartite Analysis}\label{sec:cxg}

We address the issues of construal by proposing a \textbf{bipartite analysis} that decouples the semantics signaled by the adposition from the role expected by the scene.  Essentially, we borrow from Construction Grammar \citep{fillmore-88,kay-99,goldberg-06} the notion that semantic contributions can be made at various levels of syntactic structure, 
%and the overall meaning of a sentence comes from the combination of the various levels of constructions represented in a sentence,
beginning with the semantics contributed by the lexical items. 

Under our original single-label analysis, the full weight of semantic assignment rested on the predicate's semantic role, with the indirect assumption that the predicate selects for adpostions relevant to the assignment. Under the bipartite analysis, we assign semantics at both scene and adposition levels of meaning: we capture what the scene \emph{calls} for, henceforth \textbf{scene role} and what the adposition itself \emph{codes} for, henceforth \textbf{function}. Both labels are drawn from the supersense hierarchy (\cref{fig:hierarchy}). Allowing tokens to be annotated with both a role and a function accounts for the non-congruent adposition construals, as in \cref{ex:puccini2}.%For example:

\eenumsentence{\label{ex:puccini2}\itemsep0em 
\item \label{ex:puccini-by2} The festival focuses on the works \p{by} Puccini. \\
scene role: \psst{Creator} vs.~function:~\psst{Agent}
\item \label{ex:puccini-of2} He was an expert on the works \p{of} Puccini. \\
scene role: \psst{Creator} vs.~function:~\psst{Source}
}
Bipartite analysis recognizes that both of these sentences carry the meaning represented by the supersense \psst{Creator} at the scene level, but also allows for the construal that arises from the chosen preposition: \p{by} is assigned the function of \psst{Agent} and \p{of} is assigned the function of \psst{Source}.

Our bipartite annotation scheme does not require a syntactic parse. 
It therefore does not provide a full account of constructional compositionality. 
The scene that the PP elaborates may take a variety of syntactic forms; we aim to train annotators to interpret the scene without annotating its lexical/syntactic form explicitly.
%
%The bipartite analysis is not a fully constructional analysis in that it is agnostic about what syntactic form constitutes a scene. That is, the syntactic form that is assigned the scene role could be any of the potential matrix phrases that can embed an adpositional phrase, which includes both verbal and noun phrases. Furthermore, the bipartite analysis veers from Construction Grammar in that there is little to no distinction made between the semantics of the adposition and the semantics of the adpositional phrase.\nss{I would characterize the PP as linking the meaning of the preposition with its object, and projecting the prepositional relation upward. I think this is consistent with the bipartite analysis.}\jh{that's fine. still, we use the terms intercheangeably, but I guess it doesn't matter if we say that.}\ab{btw I like "bipartite" better than "bifocal"! Have gotten used to "bipartite" in last two weeks, I guess!}
%While we are building corpora with semantic labels rather than a full-fledged construction grammar, our analysis informs how such a grammar would represent the compositionality of predicates, prepositions, and objects. 
In \cref{sec:cxg_discussion}, we sketch how a compositional Construction Grammar analysis could capture the function and the scene role at different levels of structure.%, similar to constructional treatments of metaphor in previous work.
%These differences will be discussed in \cref{sec:cxg_discussion}.

\section{Applying the Bipartite Analysis} \label{sec:applying_bipartite}

In this section, we discuss some of the more productive examples of non-congruent construals in English as well as in Hindi, Korean, and Hebrew. Hereafter, we will use the notation \rf{Role}{Function} to indicate such construals. 
Adopting the ``realization'' metaphor of articulating an idea linguistically, 
this can be read as ``\psst{Role} is realized with an adposition that marks \psst{Function}.'' %introduced by an adposition in addition to the scene role.

%In light of such examples where the semantics signaled by the preposition does not match with the information a scene calls for, we argue for a bipartite analysis where these two pieces of information are two separate construals being introduced by preposition and the scene. 

%We observe a similar phenomenon crosslinguistically, and find that certain construal patterns are common crosslinguistically while others are divergent across languages%\ab{in one of the subsections below, we need to add an example where the construals introduced by comparable adpositions are different across languages--- thinking of some good examples to discuss this...---- DONE}. 
%A few examples are presented below. 

%\psst{TODO: what construal patterns are common vs. divergent across languages?}

\subsection{Emotion and Perception Construals}

Scenes of emotion and perception \citep{dirven-97,osmond-97,radden-98} provide a compelling case for the bipartite construal analysis. 
%the function of an adposition from the scene role. 
Consider the sentences involving emotion in example \cref{ex:scared}:
%\nss{are we using ``construal'' to mean ONE of the two supersenses, or the PAIR of distinct supersenses?}\ab{as i defined construal above and since have been using it, it is one of the two simultaneously present supersenses. So one construal comes from the scene and the other one from the adposition, bipartite analysis because we have two construals}\jh{hmm, I disagree. I was under the understanding, we have two layers of semantics, one construal. We talk about construals when we compare role-function pairings of shared roles, but in a given sentence, i suggest it to be stated as a single construal coming from a multiple layers of meaning. (Construal != bipartit)}\ab{In last couple of days, I think, it has changed, I do not see how I defined construal anymore. But all the text that I had been editing or adding, I had made consistent with the definition that was there till a couple of days back. and most of my changes and additions were before that, added only one major bit in section 5 since then that I had mentioned in last couple of days. hmm, need to go through the whole paper to make sure whatever definition we are following now, the text is consistent with that! Jena- do you want to take a stab at that since I think the current description of construal is yours!}
%
\eenumsentence{\label{ex:scared}
\itemsep0em 
\item \label{ex:bear} %[Emotion]
I was scared \p{by} the bear. \\ \rf{Stimulus}{Causer}
%\item \label{ex:salary} %[Emotion] 
%I’m unhappy \p{about} my salary: \psst{Stimulus} vs.~\psst{Topic}
\item \label{ex:job} %[Emotion] 
%I was scared \p{about} losing my job. 
I was scared \p{about} getting my ears pierced. \\ \rf{Stimulus}{Topic} %\rf{Stimulus}{Topic}\ab{I don't think "about my new job" is the STIMULUS here!}\nss{perhaps better: ``I was scared about getting my ears pierced'' (found via Google search)}
%
% \item \label{ex:crowd} [Emotion] the anger \p{of} the crowd: \psst{Experiencer} vs.~\psst{Possessor}
% \item \label{ex:wait} [Emotion] The wait was unbearable \p{for} me / annoying \p{to} me.
% \psst{Experiencer} vs.~\psst{Recipient}
}
%

%For scenes involving emotion or perception, the argument that triggers the stated emotion or feeling is considered to be the stimulus. However, labeling the preposition with the scene role of \psst{Stimulus} in examples in \cref{ex:scared} does not seem sufficient for an accurate interpretation of these sentences. 
Comparing examples \cref{ex:bear} and \cref{ex:job}, we notice that there are two different types of stimuli represented in otherwise semantically parallel sentences. The preposition \p{by} gives the impression that the stimulus is responsible for triggering an instinctive fear reflex (i.e., \psst{Causer}), while \p{about} portrays the thing feared as the content or \psst{Topic} of thought.\footnote{Interestingly, ``scared \p{about}'' seems to require an explicit or metonymic event/situation as the complement. Thus, ``scared \p{about} the bear'' would be felicitous to describe apprehension about some mischief that the bear might get up to. It would be less than felicitous to describe a hiker's reaction upon being surprised by a bear.}

In some languages, the experiencer can be conceptualized as a recipient of the emotion or feeling,
thus licensing dative marking.\footnote{English displays this to a limited extent: ``It feels/seems/looks perfect \p{to} me.''} 
In the Hebrew example \cref{ex:hot_to_me}, the experiencer of bodily perception is marked with the dative preposition \p{l(e)-}  \citep{berman-82}. %, suggesting that it is interpreted as the recipient of the perception. 
Similarly, in Hindi, the dative postpostion \p{-ko} marks an experiencer in \cref{ex:seems_hot_to_me}.
%Hindi and Korean, notice the use of dative postpositions \p{-ko} in \cref{ex:seems_hot_to_me} and \p{-eykey} in \cref{ex:good_to_men}.

\eenumsentence{\label{ex:dative-experiencers}\itemsep0em 
\item \label{ex:hot_to_me} [Hebrew]: \rf{Experiencer}{Recipient}\\
Koev \p{l}-i ha-rosh \\
Hurts \p{DAT}-me the-head\\
``My head hurts.''

% xam \p{l}-i \\
% hot DAT-me \\
% ``I'm hot.''
\item \label{ex:seems_hot_to_me} [Hindi]: \rf{Experiencer}{Recipient}\\
mujh-\p{ko} garmii lag rahii hai \\
I-\p{DAT} heat feel PROG PRES \\
``I'm feeling hot.''

% \item \label{ex:good_to_men} [Korean]: %토마토가 남자에게 좋다 한다\\
% \rf{Experiencer}{Recipient}\\
% tomato-ga namca-\p{eykey} cohta-hanta.\\
% tomato-NOM  men-DAT good-EVID.\\
% ``(It is said) tomatoes are good for men.''\ab{so this one is not the experiencer-recipient construals example. If we can find an example with psych verbs with dative subjects, that would be good!}
}

Contrast this with examples where scene role and adposition function are congruent: 

\eenumsentence{
\itemsep0em 
\item \label{ex:easy1} I ate dinner \p{at} 7:00: \rf{Time}{Time}
\item \label{ex:easy2} Let’s talk \p{about} our business plan: \rf{Topic}{Topic}
%\item \label{ex:easy3} The book is \p{on} the table: \psst{Location}
%\item Russia fought \p{against} China: \psst{Opponent}
}
%
%The scene roles assigned in examples \cref{ex:easy1,ex:easy2} indicate the \psst{Time} of the eating event and \psst{Topic} of a communication event, respectively. Given these scene roles, the function of the preposition is ``natural'' for expressing these relations. 
In \cref{ex:easy1} and \cref{ex:easy2}, the preposition is prototypical for the given scene role and its function directly identifies the scene role.
Because the semantics of the role and function are congruent, these cases do not exhibit the extra layer of construal seen in \cref{ex:scared} and \cref{ex:dative-experiencers}.\footnote{One might object that most or all adpositions impose a spatial construal---and thus, \cref{ex:easy1} should be annotated as \rf{Time}{Location}. We do not discount the possibility that such a metaphor can be cognitively active in speakers using temporal adpositions; in fact, there is considerable evidence that time-as-space metaphors are cross-linguistically pervasive and productive \citep{lakoff-80,nunez-06,casasanto-08}. However, we do not see much practical benefit to annotating temporal \p{at} or topical \p{about} as spatial.} In essence, the bipartite analysis help capture the construals that characterize the less prototypical scene role and function pairings.

%\nss{discussion of emotion: \citep{dirven-97,osmond-97,radden-98}}

% maybe we should probably bring in Jackendoff 1990 discussion of two-tier analysis. It is similar to the two-tier analysis in that we recognize that an argument/adjunct (constituent?) can be analyzed for two separate semantic roles given the interpretation of the predicate. thematic and action tiers.

\subsection{Professional Associate Construals}

%Scenes involving professional associate (\psst{Associate}), previously called \psst{ProfessionalAspect} \citep{schneider-15,schneider-16}, provide another construction type where a bipartite analysis is useful in accounting for the separate construal presented by the adpositions. \psst{ProfessionalAspect} was a subtype of the supersense \psst{Undergoer} in the supersense hierarchy and was designed to mark adpositions that expressed an entity with which an individual was formally associated in a social or professional capacity (e.g., via employment or membership). While this definition generally applies to all instances in our corpus labeled with this supersense, we find that the label is semantically overloaded. Without the bipartite analysis, addressing semantic varieties in this category, as illustrated below, would call for additional subtypes under this supersense category.

Our online reviews corpus \citep{schneider-16} shows that, at least in English, professional relationships (especially employer--employee and business--client ones) are fertile ground for alternating preposition construals. The following were among the examples tagged as \psst{ProfessionalAspect}:\longversion{\footnote{We are considering replacing \psst{ProfessionalAspect} with a broader category called \psst{SocialRel} that would additionally encompass kinship and other relations between persons.}} %and moving it under \psst{Configuration}, as it is used for stative scenes (often relations between nominals).}

\eenumsentence{\itemsep0em 
\item \label{ex:associate_for}%I am a music junkie that grew up in the 80's , and 
My dad worked \p{for} a record label in the 1960's.\\
\rf{ProfessionalAspect}{Beneficiary}

%\item \label{ex:associate_with}I will NEVER do business \p{with} Sun Toyota again.\\
%\rf{ProfessionalAspect}{Co-Agent}

\item \label{ex:associate_at}Dr. Strzalka \p{at} Flagship CVTS is not a good doctor.\\
\rf{ProfessionalAspect}{Location}

\item \label{ex:associate_from} Nigel \p{from} Nidd Design has always been great! \\ \rf{ProfessionalAspect}{Source}

%provided a first class service. 

\finalversion{\item \label{ex:associate_of} the owners and employees \p{of} this store \\ \rf{ProfessionalAspect}{Possessor}}
}
All of these construals are \emph{motivated} in that they highlight an aspect of prototypical professional relationships: e.g., an employee's work prototypically takes place at the business location (hence ``work \p{at}''), though this is not a strict condition for using ``work \p{at}''---the meaning of \p{at} has been extended from the prototype. Likewise, the pattern ``\emph{person} \{\p{at}, \p{from}, \p{of}\} 
%\p{with}
 \emph{organization}'' has been conventionalized to signify employment or similar institutional-belonging relationships. %\finalversion{\footnote{For ``Nigel \p{of} Nidd Design,'' we deem ``Nigel \p{from} Nidd Design'' a closer paraphrase than ``Nidd Design's Nigel,'' and would thus choose \psst{Source} as the function over \psst{Possessor}---in contrast to \cref{ex:associate_of}.
Bipartite analysis equips us with the ability to use the existing labels like \psst{Beneficiary} or \psst{Co-Agent} to deal with the overloading of the \psst{ProfessionalAspect} label, instead of forcing a difficult decision or creating several additional categories. This analysis also accounts for similar construals presented %simplifies the process of recognizing other role-function combinations represented 
by adpositions in other languages. For example, the overlap of \psst{ProfessionalAspect} with \psst{Source}, as seen in English example \cref{ex:associate_from}, occurs in Hindi and Korean as well. %Additionally, in Hindi and Korean, the \psst{Associate} role is often construed as \psst{Source}. 

\subsection{Static vs.~Dynamic Construals}\label{sec:motion}

Another source of difficulty in the original annotation came 
from caused-motion verbs like \emph{put}, which takes a PP indicating part of a path. 
Sometimes the preposition lexically marks a source or goal, e.g., \p{into}, \p{onto}, or \p{out of} \cref{ex:put_into}.
Often, however, the preposition is prototypically locative, e.g.,~\p{in} or \p{on} \cref{ex:put_on}, though the object of the preposition is interpreted as a destination, equivalent to the use of \p{into} or \p{onto}, respectively. 
This locative-as-destination construal is highly productive, so analyzing \p{on} as polysemous between \psst{Location} and \psst{Destination} does not capture the regularity.
The PP is sometimes analyzed as a resultative phrase \citep{goldberg-06}. 
In our terms, we simply say that the scene calls for a \psst{Destination}, 
but the preposition codes for a \psst{Location}:
\eenumsentence{\label{ex:cynthia}\itemsep0em 
\item \label{ex:put_into} Cynthia put her things \p{into} a box.\\ \rf{Destination}{Destination} %destination so that we don't have to explain why there isn't an EndLocus in the current hiearchy and why Destiation, which is on the hierarchy isn't used?
\item \label{ex:put_on} Cynthia put her things \p{on} her bed.\\ \rf{Destination}{Location}
}
Thus, we avoid listing the preposition with multiple lexical functions for this regular phenomenon.

The opposite problem occurs with fictive motion \citep{talmy-96}: a path PP, and sometimes a motion verb, construe a static scene as dynamic:

\enumsentence{\label{ex:river_runs} A road runs \p{through} my property. \rf{Location}{Path}}
Rather than forcing annotators to side with the dynamic construal effected by the language, versus the static nature of the actual scene, we represent both: the scene role is \psst{Location} (static) and the preposition function is \psst{Path} (dynamic).

\section{Challenges and Opportunities}\label{sec:challenges}

%In this section, we discuss current challenges, and evaluate potential risks and benefits as we consider the adaptation of the bipartite analysis to the preposition supersense project. 

The added representational complexity of the bipartite analysis seems justified to account for many of the phenomena discussed above, especially as the project grows to include more languages. But is the complexity worth it on balance? We consider some of the tradeoffs below.

\subsection{Challenges in Function Assignment} \label{sec:fxn_challenges}
%With some of adposition or case marker usages, converging on the identity of a function of the adposition or case marker %as used in the sentences 
%presents itself as a challenge in the bipartite analysis. While separating the construal corresponding to the scene role from the construal arising from the adposition or case marker, i.e., the construal corresponding to the function, does lead to a better interpretation in cases where the two construals do not match. But, 
We %have 
encountered several examples in which function labels are difficult to identify. %quantify. 
Consider the following paraphrases:

\eenumsentence{\itemsep0em 
\item \label{ex:run-eso} [Korean]: %나는 철수\textbf{와} 만났다\\
\rf{Location}{Location}\\
Cheolsu-nun undongcang-\p{eyse} tallyessta.\\
Cheolsu-NOM schoolyard-\p{at} ran.\\
``Cheolsu ran in the schoolyard.''
\item \label{ex:run-ul} [Korean]: %나는 철수\textbf{를} 만났다\\
\rf{Location}{?}\\
Cheolsu-nun undongcang-\p{ul} tallyessta. \\
Cheolsu-NOM schoolyard-\p{ACC} ran.\\
``Cheolsu ran in the schoolyard.''
% \item \label{ex:meet-wa} [Korean]: %나는 철수\textbf{와} 만났다\\
% \rf{Accompanier}{Accompanier}\\
% na-nun Chelswu-\p{wa} manna-ss-ta.\\
% I-NOM Chelswu-\p{with} meet.\\
% "I met with Chelswu."
}
%
%The verb \textit{run} in Korean calls for the role of \psst{Location}, and is often realized with the dedicated locative pospostion \p{-eyse} but not always. Here we have two semantically equivalent sentences, which translate to ``Cheolsu ran in the schoolyard''.  
In \cref{ex:run-eso}, ``schoolyard'' is accompanied by a postposition \p{-eyse} (comparable to English \p{at}), which marks it as the location of running. This is the unmarked choice. On the other hand, in sentence \cref{ex:run-ul}, the noun is paired with the accusative marker \p{-ul}, the marked choice. The use of \p{-ul} evokes a special construal: it indicates that the schoolyard is more than just a backdrop of the running act and that it is a location that Cheolsu mindfully chose as the place of action. Additionally, marking the location with the accusative marker, pragmatically, brings focus to the noun (i.e., he ran in a schoolyard as opposed to anywhere else). 
%%% Next paragraph replaces the text commented just below it, for space
\longversion{ %\nss{clarify/add examples}The construal given the accusative marker is found both in Korean and Hindi.
%is fairly productive in Korean. 
%Construals that arise from the accusative marker in general are fairly productive in both Korean and Hindi.\ab{I wouldn't say they are fairly productive in Hindi}
Such construals are not limited to locations, but may also include other scene roles such as \psst{Goal} and \psst{Accompanier}, in alternation with postpositions that can express those functions.
%Additionally, s
Since accusative case markers generally serve syntactic functions over semantic ones, it may be difficult to identify a semantic function the accusative marker carries.

A similar phenomenon can be found in Hindi:

\eenumsentence{
  \item\label{ex:library-bare} [Hindi]: bare NP as \psst{Destination} \\
  maiN library jaa rahii thii \\
  I library go PROG PST \\
  ``I was going to the library.''

  \item\label{ex:library-acc} [Hindi]: \rf{Destination}{?} \\
  maiN library-\p{ko} jaa rahii thii \\
  I library-\p{ACC} go PROG PST \\
  ``I was going to the LIBRARY.'' [more emphasis on the library]
}
This suggests that, apart from spatiotemporal relations and semantic roles, 
adpositions can mark \textbf{information structural} properties 
for which we would need a separate inventory of labels.
}

In some idiomatic predicate--argument combinations, the semantic motivation for the preposition may not be clear \cref{ex:idiomatic}. %For example:

\eenumsentence{\label{ex:idiomatic}\itemsep0em 
\item \label{ex:listen_to} Listen \p{to} the violin! \rf{Stimulus}{?}
\item \label{ex:proud_of} What are you proudest \p{of}? \rf{Stimulus}{?}
\item \label{ex:unhappy_with} I was unhappy \p{with} my meal. \rf{Stimulus}{?}
\item \label{ex:interested_in} Are you interested \p{in} politics?
\rf{Topic}{?}
}
While the scene role in \cref{ex:listen_to} and \cref{ex:proud_of} is clearly \psst{Stimulus}, the function is less clear. Is the object of attention construed (metaphorically) as a \psst{Goal} in \cref{ex:listen_to}, and the cause for pride as a \psst{Source} in \cref{ex:proud_of}? Or are \p{to} and \p{of} semantically empty argument-markers for these predicates (cf.~the ``case prepositions'' of \citealp{rauh-93})? We do not treat either combination as an unanalyzable multiword expression because the ordinary meaning of the predicate is very much present. 
\cref{ex:unhappy_with} and \cref{ex:interested_in} are similarly fraught.
But as we look at more data, we will entertain the possibility that the function can be null to indicate a marker which contributes no lexical semantics.

% \eenumsentence{
% \itemsep0em 
% \item \label{ex:listen} Listen \p{to/for} the violin! \rf{Stimulus}{?}
% \item \label{ex:look} Look \p{at/for} the violin! \rf{Stimulus}{?}
% }

% The prepositional phrases in \cref{ex:listen,ex:look} serve the role of stimuli to their respective verbs of sensory experience. What is unclear, however, is what specific functions these prepositions have in the sentences. It is fairly clear that these prepositions do give rise to different construals: the distinction between \p{to} vs.~\p{for} in \cref{ex:listen} and \p{at} vs.~\p{for} in \cref{ex:look} is that the latter prepositions of the two pairs indicate the act of searching, which is missing from the former prepositions in the two pairs. It is, however, difficult to pinpoint the exact function these former prepositions carry. %It is not clear what specific function the former prepositions in the two pairs carry. %While the existence of construal suggests that there is a functional role played by the preposition, converging at what it is, is a problem to which we are still working towards a satisfactory solution.

\subsection{The Annotation Process}

%%%% removing for space 
% Another concern regarding the adoption of the bipartite analysis would present for preposition supersense annotation is in regards to the practicality of the described approach to the constraints and needs of an annotation project. As the preposition supersense project was specifically designed with the need to produce annotated datasets for training and evaluating natural language understanding systems, it an important concern.
% %In our current annotation practice where a single label is assigned to an adposition, the task of annotating involves identifying the role of the adposition given the context. 
% Adopting the bipartite analysis splits the annotation process into two subtasks: the annotation of the scene role based on the context of the adposition and the annotation of the function label based on the prototypical senses of the adposition. While the task of annotating scene roles are not too different in practice from what the supersense annotators have been trained to do, the task of annotating function labels poses a concern that needs to be addressed.

Annotators are generally capable of interpreting meaning in a given context.
%, which is what makes projects like FrameNet and PropBank annotation tasks.  
However, it might be difficult to train annotators to develop intuitions about adposition functions, which reflect prototypical meanings contributed by the lexical item that may not be literally applicable. These distinctions may be too subtle to annotate reliably.
%However, if they are being asked to speculate about the nearest prototypical sense of an adposition---essentially, to make a linguistic analysis about the contribution of the adposition, we may be requiring from the annotators the familiarity with and the ability to discern distinctions that are relevant for theories of semantic decomposition but might not be as obvious to general native speaking population. 
As we are approaching this project with the goal of producing annotated datasets for training and evaluating natural language understanding systems, it is an important concern.

%Once the linguistics issues as those discussed in \cref{sec:fxn_challenges} are addressed, 
We are currently planning pilot annotation studies to ascertain (i)~the prevalence of the role vs.~function mismatches, and (ii)~annotator agreement on such instances.
Enshrining role--function pairs in the lexicon may facilitate inter-annotator consistency: our experience thus far is that annotators benefit greatly from examples illustrating the possible supersenses that can be assigned to a preposition.
If initial pilots are successful, we would then need to decide whether to annotate the role and function together or in separate stages. 
Because the function reflects one of the adposition's prototypical senses, it may often be deterministic given the adposition and scene role, in which case we could focus annotators' efforts on the scene roles.
Existing annotations for lexical resources such as PropBank \citep{palmer-05}, VerbNet \citep{kipper-08}, and FrameNet \citep{framenet} might go a long way toward disambiguating the scene role, limiting the effort required from annotators.

\subsection{Linguistic Utility of Annotated Data}

Assuming the above theoretical and practical concerns are surmountable, annotated corpora would facilitate empirical studies of the nature and limits of adposition/case construal within and across languages. For example: Is it the case that some of the supersense labels can only serve as scene roles, or only as functions? (A hypothesis is that \psst{Participant} subtypes tend to be limited to scene roles, but this needs to be examined empirically.) 
Which role--function pairs are attested in particular languages, and are any universal?
Thus far we have seen that certain scene roles, such as \psst{Experiencer}, \psst{Stimulus}, and \psst{ProfessionalAspect}, invite many different adposition construals---is this universally true?
As adpositions are notoriously difficult for second language learners, would it help to explain which construals do and do not transfer from the first language to the second language?

\subsection{Modifying the Supersense Hierarchy}

The bipartite analysis may allow us to trade more complexity at the token level for less complexity in the label set.
%While the necessity of considering two supersense labels instead of one adds complexity to the scheme at the token level, we surmise that the bipartite analysis will help with the hierarchy in several ways. Because the construal phenomenon is sufficiently productive, it has become unwieldy to add new multiple inheritance nodes for each combination of scene role and function. 
As discussed in \cref{sec:construal}, separating the scene role and function levels of annotation will more adequately capture construal phenomena without forcing an arbitrary choice between two labels or introducing further complexity into the hierarchy. %As a consequence, this approach shows promise for reducing the complexity of the supersense label inventory by allowing us to employ existing labels rather than having to posit a new label for each available construal. 
%
%Additionally, as discussed in the paper, the bipartite analysis will help circumvent some of the issues presented by cross-linguistic efforts without necessitating the addition of a large number of new labels.  This will be especially important for language like Korean for which a small set of pilot annotation seem to indicate construals are highly pervasive.
%steels-12
%Finally, 
In fact, we hope to \emph{simplify} our current supersense hierarchy, especially by removing labels with multiple inheritance for usages that can be accounted for with the bipartite analysis instead. Candidates include \psst{Contour} (inheriting from \psst{Path} and \psst{Manner}; e.g., ``The fly flew \p{in} zig-zags'') and \psst{Transit} (inheriting from \psst{Via} and \psst{Location}; e.g., ``We traveled \p{by} bus''). We may also 
%move \psst{State}, \psst{StartState}, and \psst{EndState} under %\psst{Circumstance}, and 
collapse the pairs \psst{Locus}/\psst{Location}, \psst{Source}/\psst{InitialLocation}, and \psst{Goal}/\psst{Destination}. %Bipartite analysis opens up the opportunity to reevaluate instances belonging to such supersenses to identify and to remove those categories that were created for the sole purpose of capturing some of the finer grained construal distinctions in English. 
%As well as reducing the number of labels for annotators to consider, 
A simpler hierarchy of supersenses will serve to reduce the number of labels for annotators to consider during the annotation process and also help improve automatic methods by reducing sparsity of labels in the data.

\section{Discussion of Constructional Analysis} \label{sec:cxg_discussion}

We have focused on developing a \emph{broad-coverage} annotation scheme for adpositional semantics, 
and our proposal requires no more than two categorical labels per adposition token\longversion{ (but see below)}.
Although our current approach falls short of a full constructional derivation of the form--meaning correspondences that comprise a sentence and the interpretation that results, we believe our approach could inform such an analysis.

Construction Grammar formalisms that support full-sentence analyses include Embodied Construction Grammar \citep{bergen-05},
%,feldman-09}, % removed chang-08
Fluid Construction Grammar \citep{steels-11}, %steels-12
and Sign-Based Construction Grammar \citep{%
%sag-07,michaelis-09,sag-12,
boas-12}.
Without tying ourselves to any one of these, we observe at a high level that the lexical semantic contribution of the adposition (the function) can be distinguished from the role of a governing predicate or scene by assigning these meanings to different stages of the derivation. 
E.g., in ``care \p{about} the strategy,'' the adposition and PP could express a \psst{Topic} figure-ground relation whose ground is the meaning of ``the strategy''; ``care'' could evoke a semantic frame with a \psst{Stimulus} role; and an argument structure construction could link the ground of the figure-ground relation with the \psst{Stimulus} role. 
If the \rf{Stimulus}{Topic} construal with \p{about} is sufficiently productive, the generalization could be formalized via an argument-structure construction with a verb slot limited to verbs of (say) emotion and a PP headed by topical \p{about}.
%[[care]$_V^\text{Caring}$ [of]$_P^\text{\psst{Possessor}} Bipashaa]NP

There are complications which we are not yet prepared to fully address. 
First, if the PP is not governed by a predicate which provides the roles---such as a verb or eventive/relational noun---the preposition may need to evoke a meaning more specific than our labels. 
E.g., for ``children \p{in} pajamas'' and ``woman \p{in} black,'' \p{in} may be taken to evoke the semantics of wearing clothing.\footnote{Indeed, this is the position adopted by version~1.7 of FrameNet, where \p{in} is listed as a lexical unit of the \textsc{Wearing} frame\longversion{ (\url{https://framenet2.icsi.berkeley.edu/fnReports/data/frame/Wearing.xml})}.} 
The label set we use for broad-coverage annotation is, of course, vaguer, and would simply specify \psst{Attribute} for the clothing sense of \p{in}. 
Copular constructions raise similar issues. 
Consider ``It is \p{up to} you to decide,'' meaning that deciding is the addressee's responsibility: this idiomatic sense of \p{up to} is closer to a semantic predicate than to a semantic role or figure-ground relation.

% Some prepositions carrying non-spatiotemporal semantics are associated with highly productive PP adjunct constructions. For example, in English, benefactive \p{for}-PPs can apply to almost any activity, so long as it can be understood as benefiting some party:
% \eenumsentence{\label{ex:benefactive}
% \item I'll bake a cake \p{for} you.
% \item I work hard \p{for} my family.
% %\item They will perform \p{for} a large audience.
% \item \p{For} their corporate clients, they lobby against government regulations.
% }
% The preposition \p{for} is one of the prototypical markers for targets of an action, namely, \psst{Goal} (e.g., ``studying \p{for} an exam''\nss{I think this is \psst{Purpose}}) and \psst{Recipient} (e.g., ``a birthday present \p{for} me''). It seems likely that the benefactive construction emerged by grammaticalization of a metaphor in which assistance is `targeted' at someone (the \psst{Beneficiary}). 
% Is this metaphor still conceptually active in the beneficative \p{for}-PP construction? If so, it would motivate annotating the examples in \cref{ex:benefactive} as \rf{Beneficiary}{Goal} (or \rf{Beneficiary}{Recipient}).
% However, this policy may lead down the path of treating \emph{all} non-spatiotemporal meanings with the bipartite analysis. To keep the annotation task tractable, it may be best to stick to a single label for productive adjunct constructions such as benefactive \p{for}-PPs.

\longversion{
In rare instances, we are tempted to annotate a chain of extensions from a prototypical function of a preposition, which we term \textbf{multiple construal}. For instance:
\eenumsentence{\label{ex:dbl}
\item\label{ex:yell_at} Bob's boss yelled \p{at} him for his mistake.\\
\rff{Recipient}{Beneficiary}{Goal}
\item\label{ex:angry_at} Bob's boss was angry \p{at} him for his mistake.\\
\rff{Stimulus}{Beneficiary}{Goal}
\item\label{ex:involved_in} I was involved \p{in} the project.\\ \rff{Theme}{Superset}{Location}}

``Yelled \p{at}'' in \cref{ex:yell_at} is a communicative action whose addressee (\psst{Recipient}) is also a target of the negative emotion (\rf{Beneficiary}{Goal}: compare the use of \p{at} in ``shoot \p{at} the target'').
\cref{ex:angry_at} is similar, except ``angry'' focuses on the emotion itself, which Bob is understood to have evoked in his boss.

With regard to \cref{ex:involved_in}, the item ``involved \p{in}'' has become fossilized, with \p{in} marking an underspecified noncausal participant (hence, \psst{Theme} as the scene role).
At the same time, one can understand the \p{in} here as motivated by the member-of-set sense (cf.~``I am \p{in} the group''), which would be labeled \rf{Superset}{Location} because it conceptualizes membership in terms of containment. A similar logic would apply to ``people \p{in} the company'': \rff{ProfessionalAspect}{Superset}{Location}.
Effectively, the multiple construal analysis claims that multiple steps of extending a preposition's prototypical meaning remain conceptually available when understanding an instance of its use.
% \footnote{In the formal semantics literature, 
% a strict separation of semantic structure and conceptual structure known as the Two-Level Model has been proposed as an economical account of polysemy: 
% a polysemous word's semantic form is taken to be a schematic generalization that is shared among all of its related usages, 
% and the different usages are attributed to an ``essentially non-linguistic, conceptual level of interpretation'' \citep[p.~3]{taylor-95}. 
% While the use of multiple levels or dimensions of meaning representation is broadly in keeping with our approach, 
% it is not clear that a firm separation between semantic structure and conceptual structure can be maintained, 
% especially in light of multiple construal.\nss{TODO: revisit}} 
That said, we are not convinced that this logic could be applied reliably by annotators, and thus may simplify the usages in \cref{ex:dbl} to just the first and second or the first and third labels.
}

% \jh{potential alternate to \ref{ex:scared_hin}}

% Another instance where a double construal analysis might be relevant is exemplified in Hindi example \ref{ex:scared_hin2}. 

%  \enumsentence{ \label{ex:scared_hin2}
%  [Hindi]:  \rff{Stimulus}{Topic}{Causer}\\
%  maiN     kaan chidwaane	\p{se}     Dar	gayaa thaa \\
%  I		   ear to.pierce      \p{from}     afraid	PERF PST \\
%  ``I got afraid of ear piercing.''
%  }

% This sentence is the most natural translation of \ref{ex:job} in that the postpositional phrase indicates the content of fear (i.e., \rf{Stimulus}{Topic}) rather than an instinctive fear reflex as indicated by the English example \ref{ex:bear}. At the same time, the postposition used here is \p{se}, giving the impression that the stimulus is responsible for triggering a fear reflex and suggesting a \psst{Causer} construal (i.e., \rf{Stimulus}{Causer}). Double construal analysis would allow for both of levels of construal to be represented in the annotation.

% \jh{end of potential alternate}

Finally, metaphoric scenes \citep{lakoff-80} raise a whole host of issues. In \cref{ex:put_metaphor}, the locative-as-destination construal (\cref{sec:motion}) is layered with the states-are-locations metaphor.
In bipartite analysis, we annotate the scene in terms of the governing predicate's \textbf{target domain}, and the adposition function in terms of the \textbf{source domain}:
\enumsentence{\label{ex:put_metaphor} 
The election news put him in a very bad mood.\\ \rf{EndState}{Location}
}
A constructional analysis could capture both source domains and both target domains---i.e., \psst{Location}, \psst{State}, \psst{Destination}, and \psst{EndState}--perhaps by assigning source domain meanings to lexical constructions and target domain meanings to their mother phrases.\footnote{\Citet{dodge-14} similarly analyze ``attacking poverty'' by assigning the source domain meaning (i.e., a literal attack) to ``attacking'' and the target domain meaning (`solving a social problem') to the argument structure construction.}

% outstanding questions when it comes to cxg:

% - preposition vs. prepositional phrase

% - do the nominal head constructions behave different from that of verb phrase level ones? (question of what is a predicate)

% \nss{TODO: address ``what is a scene?'' earlier in the paper} 
% \jh{gave a 1 line def in paragraph 2 in 4.1}

%\nss{TODO: cite Metaphors We Live By, e.g. States are Locations, somewhere.}

\section{Conclusion}\label{sec:conclusion}
We have considered the semantics of adpositions and case markers in English and a few other languages with the goal of revising a broad-coverage annotation scheme used in previous work. We pointed out situations where a single supersense did not fully characterize the interaction between the adposition and the scene elaborated by the PP. In an attempt to tease apart the semantics contributed specifically by the adposition from the semantics coming from elsewhere, we proposed a bipartite construal analysis.
%an analysis based on separate construals introduced by adpositions and by scenes\ab{this statement would need to be changed to be consistent with the current construal definition.}. 
Though many details remain to be worked out, we are optimistic that our bipartite analysis will ultimately improve broad-coverage annotations as well as constructional analyses of adposition behavior.

\section*{Acknowledgments}

We thank the rest of our CARMLS team---Martha Palmer, Ken Litkowski, Katie Conger, and Meredith Green---% 
for participating in weekly discussions of adposition semantics; 
Michael Ellsworth for an insightful perspective on construal, 
Paul Portner for a helpful clarification regarding approaches to conceptualization in the literature, 
and anonymous reviewers for their thoughtful comments.

\bibliographystyle{aaai}
\begingroup
	%\small
	\setlength{\bibsep}{2pt}
  \bibliography{prepconstrual.bib}
\endgroup

\end{document}